\documentclass[sn-mathphys-ay]{sn-jnl}
\usepackage{graphicx}%
\usepackage{multirow}%
\usepackage{amsmath,amssymb,amsfonts}%
\usepackage{amsthm}%
\usepackage{mathrsfs}%
\usepackage{xcolor}%
\usepackage{textcomp}%
\usepackage{manyfoot}%
\usepackage{booktabs}%
\usepackage{algorithm}%
\usepackage{algorithmicx}%
\usepackage{algpseudocode}%
\usepackage{listings}%
\usepackage{newtxtext,newtxmath}
\usepackage{orcidlink}
\usepackage{subcaption}

\begin{document}

\title{Revisiting Expected Possession Value in Football: Introducing a Benchmark, U-Net Architecture, and Reward and Risk for Passes}

\author*[1,2]{\fnm{Thijs} \sur{Overmeer}\orcidlink{0009-0003-9108-1909}}\email{t.overmeer@tue.nl}
\author[3]{\fnm{Tim} \sur{Janssen}\orcidlink{0000-0002-8050-1176}}
\author[1,2]{\fnm{Wim P.M.} \sur{Nuijten}\orcidlink{0000-0003-0351-2768}}

\affil[1]{\orgdiv{Eindhoven AI Systems Institute}, \orgname{Eindhoven University of Technology}, \orgaddress{ \city{Eindhoven}, \country{the Netherlands}}}
\affil[2]{\orgdiv{Department of Mathematics and Computer Science}, \orgname{Eindhoven University of Technology}, \orgaddress{ \city{Eindhoven}, \country{the Netherlands}}}
\affil[3]{\orgname{Royal Dutch Football Association (KNVB)}, \orgaddress{ \city{Zeist}, \country{the Netherlands}}}

\date{\today}

\abstract{
This paper introduces the first Expected Possession Value (EPV) benchmark and a new and improved EPV model for football. %
Through the introduction of the OJN-Pass-EPV benchmark, we present a novel method to quantitatively assess the quality of EPV models by using pairs of game states with given relative EPVs. %
Next, we attempt to replicate the results of \cite{Fernández2021} using a dataset containing Dutch Eredivisie and World Cup matches. Following our failure to do so, we propose a new architecture based on U-net-type convolutional neural networks, achieving good results in model loss and Expected Calibration Error. %
Finally, we present an improved pass model that incorporates ball height and contains a new dual-component pass value model that analyzes reward and risk. %
The resulting EPV model correctly identifies the higher value state in 78\% of the game state pairs in the OJN-Pass-EPV benchmark, demonstrating its ability to accurately assess goal-scoring potential. %
Our findings can help assess the quality of EPV models, improve EPV predictions, help assess potential reward and risk of passing decisions, and improve player and team performance.}

\begin{titlepage}
    \maketitle

\end{titlepage}

 \newpage

 \section{Introduction}

In recent years, football analytics has become increasingly important for gaining a competitive edge. This paper focuses on a specific metric in the expanding realm of football data analysis, \textit{Expected Possession Value} (EPV) (\cite{fernandez2019decomposing}), which quantifies the probability of scoring or conceding a goal given the current \textit{game state}.

The first research question we address is how to evaluate the quality of any EPV model. This is a general question for the research field and to start answering it we introduce the OJN-Pass-EPV benchmark consisting of pairs of game states, where an expert panel determines the game state that they deem to be more dangerous. The OJN-Pass-EPV benchmark focuses on the relative \textit{pass} value of game states and as such can be used to evaluate \textit{pass models}.

Our second research question is whether we can replicate the results of \cite{Fernández2021} using a dataset from the Dutch Eredivisie and the 2022 World Cup before attempting to improve upon their model. Our attempts reveal discrepancies in model parameters and issues related to the vanishing gradient problem when employing a similar network structure for the pass model. To overcome this, we introduce a more classical U-net-type convolutional neural network structure. We test our model on the two datasets, delving into a relatively unexplored area of assessing the adaptability and robustness of a model across different levels of football competition.

Finally, we focus on \textit{pass EPV}, splitting the \textit{pass value} into two distinct components: \textit{reward} and \textit{risk}, for both successful and unsuccessful passes. This approach allows for a more detailed assessment of each pass.
By separately quantifying the reward (the potential positive impact of a pass) and the risk (the potential negative impact of a pass), our model offers an improved view of player decision making. We furthermore include ball height as a new feature, as such adding the crucial vertical dimension of play and allowing to distinguish between aerial and ground passes.

Our main contributions are:
\begin{itemize}
    \item The OJN-Pass-EPV benchmark, consisting of pairs of game states with given relative EPVs, can be used to assess the quality of any pass EPV model.
    \item A new U-net-type convolutional neural network architecture for the evaluation of pass EPV.
    \item Assessment of the adaptability and robustness of our model in different levels of football competition.
    \item The splitting of the pass value into reward and risk for both successful and unsuccessful passes.
    \item The inclusion of ball height as a new feature.
\end{itemize}

The following sections review the relevant literature, detail our methodology, present the results, and discuss the implications of our findings for football analytics and strategy.
\newpage
 
 \section{Literature Review} \label{literature_review}

Assessing the probability of a ball possession leading to a goal is fundamental in football analytics. Several methods have been suggested to measure this potential, each with distinct advantages and disadvantages.

Possession-based metrics such as VAEP \citep{VAEP2019} and xThreat (xT) \citep{singh} rely solely on \textit{event data}, which pertain to on-ball actions, to quantify the added value of actions in terms of goal-scoring probability.

Alternative metrics, such as Dangerousity \citep{Link2016}, the value of passes model \citep{Power2017}, and expected pass \citep{anzer2022expected}, utilize \textit{tracking data}, capturing the positions of all players on the pitch multiple times per second. This provides a deeper insight into game dynamics, yet often lacks the necessary interpretability for practical use. Additionally, these models often rely on simplified or partial representations of the current game state rather than integrating the full game state information, including each player's instantaneous position and velocity.

\cite{fernandez2019decomposing} introduce the Expected Possession Value (EPV) model to football as a framework for estimating the probability of a team scoring or conceding the next goal at any given moment during a possession. The model decomposes this value into the expected outcomes of three main actions: passes, ball drives, and shots. Using spatio-temporal tracking data, the model provides visually interpretable surfaces for assessing the expected value of passes, defined as the probability a goal is scored or conceded within the next 15 seconds of play, and the pass success probabilities, thereby allowing fine-grained evaluations of game states. \cite{Fernández2021} extend this framework by refining the granularity of analysis. This refinement involves explicit modeling of additional components like pass likelihood (for every location the likelihood it is the destination of a pass), dribble success evaluation, and action selection probabilities. \cite{Fernández2021} describe the results for each model, focusing on model calibration to improve interpretability and accuracy. Together, these models offer a comprehensive and interpretable tool for analyzing football dynamics, leveraging the spatial and contextual nuances captured in tracking data. Hereafter, we refer to this model as F21-EPV.



While F21-EPV represents a significant advance in football analytics, our analysis identifies and addresses the following areas of potential improvement.

\begin{enumerate}
    \item \textbf{Comparative Analysis:} 
        To our knowledge, there is currently no published method for assessing the relative value of game states to determine the more valuable state and comparing this with the output of an EPV model. We present a first step by focusing on the pass EPV model and by utilizing domain experts (see Section~\ref{benchmark_creation}).
    \item \textbf{Ball Height:}
        Previous models, including F21-EPV, do not account for the vertical dimension (height) of the ball. Our model incorporates the ball's z-axis (height), recognizing the distinct dynamics of aerial versus ground passes to improve accuracy. This addition is important, as the ball is often not at ground level during small portions of a football match. This property has received relatively little attention in football research, with the notable exception of \cite{haland2020evaluating}, who demonstrated that aerial passes have a lower success rate than ground passes.

    \item \textbf{Cross-Competition Training:}
        Most existing models are trained on data from a single competition. We train our model on multiple competitions, specifically the Dutch Eredivisie and the 2022 World Cup.
    
    \item \textbf{Model Replicability:}  
        We encountered difficulties replicating the results reported by \cite{Fernández2021} when applying their approach to our dataset. To overcome this, we adopt an architecture inspired by the conventional U-Net framework \citep{ronneberger2015unet}. 

    \item \textbf{Risk-Reward Decomposition:}
        F21-EPV provides a single EPV value for each potential pass destination. In contrast, we decompose the value of passes into separate risk and reward components, providing insights into the potential benefits and drawbacks of each pass destination.
\end{enumerate}

This research aims to improve the accuracy and interpretability of EPV models in football and to contribute to a deeper understanding of game dynamics and decision making. The OJN-Pass-EPV benchmark provides an evaluation framework for assessing the performance of pass models.
 \newpage
 \section{Methodology}

\subsection{Benchmark Creation}\label{benchmark_creation}

To enable quantitative evaluation of the performance of EPV models, we create the OJN-Pass-EPV benchmark. This benchmark consists of 50 \textit{modified} game state pairs, where we use a real game state and realistically alter aspects of it (e.g. player positions and velocities). 
The OJN-Pass-EPV benchmark focuses solely on the overall pass value of game states, making it suitable for evaluating pass models. However, the principle behind the OJN-Pass-EPV benchmark can be used to evaluate other aspects, such as \textit{dribble value}, \textit{shot value}, action selection probabilities, and ultimately, overall EPV.

\cite{davis2024methodology} emphasize that evaluating the effectiveness of deep learning models in sports is both essential and challenging due to the intricate and noisy nature of sports data and outline that one approach to do such an evaluation is to perform an analysis with the assistance of domain experts. Following that we rely on a panel of football experts, including members of the Royal Dutch Football Association (KNVB), to judge which of the two game states in each game state pair has a higher pass EPV. We focus on relative EPVs rather than absolute EPVs, as the latter are often more subjective. While acknowledging that even relative judgments can be subjective, we hypothesize that this is less likely to be a significant issue. In designing the benchmark, we select game state pairs that we expect to have widely accepted relative values. 

Details of the benchmark creation process and the complete set of game state pairs are available in the GitHub repository: 
\citep{Overmeer2025} \url{https://github.com/EAISI/OJN-EPV-benchmark}.
Note that to ensure unbiased evaluation, the game states used in the benchmark are excluded from our training, validation, and test sets.

\subsection{Initial Replication Attempts and Challenges}

In replicating the F21-EPV model, we encounter discrepancies in model parameters and the vanishing gradient problem, which hinder our ability to reproduce the reported results using our dataset. These challenges, potentially stemming from subtle implementation differences or dataset variations, led us to adjust our model's architecture. By exploring alternative activation functions, loss calculations, and layer configurations, we develop OJN-EPV, an enhanced EPV model that addresses these challenges.

\subsection{Data Collection}

The data used in this study are sourced from the KNVB, encompassing the 2021/22 and 2022/23 seasons of the Dutch Eredivisie, as well as data from the 2022 World Cup. The data consists of 10 Hz tracking data for player and ball positions, accompanied by event data. The dataset is collected and processed ensuring the alignment of both event and tracking data.

Our analysis uses data from 624 Eredivisie matches and 63 World Cup matches. This combination captures a range of performance levels and playing styles, providing a solid foundation for OJN-EPV.

\subsection{Data Preprocessing and Feature Engineering}

We transform the raw event and tracking data for integration into our models through the following steps:

\begin{itemize}
    \item \textbf{Scaling and Smoothing:} To optimize data processing, we scale the X and Y coordinates to a field representation of 103x67, resulting in a grid of 104x68 for efficient indexing in NumPy and TensorFlow. Velocities are smoothed using a Savitzky-Golay filter to reduce noise \citep{shaw_2020}.

    \item \textbf{Normalization and Cleaning:}  We standardize the data by ensuring all attacks proceed uniformly from left to right. Additionally, we remove instances of players recorded outside the pitch boundaries to improve data integrity.

    \item \textbf{Real Playing Time Calculation:} To accurately assess pass value, we calculate the actual playing time, excluding periods when the ball is out of play. This ensures that our 15-second evaluation window following each pass reflects only the active duration of the game, providing a more precise assessment of in-game actions.

    \item \textbf{Data Alignment:}  To ensure synchronicity, we align the event data with the tracking data. This ensures that each pass event is accurately reflected in the tracking data, enabling precise spatial and temporal analysis.
\end{itemize}

For the pass likelihood, pass success, and pass value models, we use the features described in \cite{Fernández2021} and additionally incorporate the z-value (height) of the ball.

\subsection{Model Architecture} \label{model_architecture_and_selection}

Our pass EPV model employs a U-Net-type convolutional neural network (CNN) architecture \citep{ronneberger2015unet}. This architecture processes game state data represented on a 104x68 grid, with the output mirroring the input dimensions.  The model comprises encoder and decoder blocks, incorporating max pooling, replication padding, attention gates, and concatenation layers.

Each encoder block features two repetitions of the following sequence: a replication padding layer, a convolutional layer with filter sizes of 16, 32, 64, 32, or 16 (depending on the block), a 5x5 filter, batch normalization, and a LeakyReLU activation function (alpha= 0.1). Decoder blocks consist of an upsampling layer, replication padding, a convolutional layer with matching filter numbers to the concatenated outputs, a 5x5 filter, batch normalization, and a LeakyReLU activation function (alpha= 0.1).

Each encoder block in the contracting path is followed by max pooling to reduce dimensionality and prevent overfitting. Max pooling is omitted in the third encoder block to preserve dimensions.  The most contracted feature maps have dimensions of 26x17.

Two decoder blocks upsample the feature maps, each followed by an attention gate, concatenation, and an encoder block. This symmetrical design captures both local and global context.

The final layer utilizes a sigmoid activation for the pass success model and softmax for the pass likelihood model. The pass value model includes a softmax layer with three classes, representing goal outcomes within a 15-second timeframe: a goal for the passing team, no goal, or a goal for the opposing team. For this model, we apply categorical cross-entropy loss to the output probabilities (e.g., [0.1, 0.8, 0.1] indicating the probabilities of a goal by the same team, no goal, and a goal by the opponent within 15 seconds, respectively) compared to the ground truth (e.g., [0, 1, 0] if no goal is scored within 15 seconds). This 15-second duration is based on Fernández et al. (2021), reflecting the average possession duration in football.

Attempts using a sigmoid activation and linear transformation for pass value, similar to \cite{Fernández2021}, encountered vanishing gradient issues.  The U-Net architecture is selected for its proven effectiveness in image segmentation and its ability to capture both local and global contextual information, which is crucial for accurate EPV predictions.

\subsection{Model Training and Evaluation}

We split the matches into training, validation and test sets using an 80-10-10 split for Eredivisie matches and a 60-20-20 split for World Cup matches. Due to the smaller size of the World Cup data set, we assigned a higher percentage of samples to the validation and test sets to enhance their statistical relevance. Table \ref{tab:passes_comparison} shows the number of pass samples for each set.

\begin{table}[ht]
\centering
\caption{Comparison of Successful and Unsuccessful Passes}
\label{tab:passes_comparison}
\begin{tabular}{lccccc}
\hline
Dataset & Total & Training & Validation & Test & \%\_success \\ \hline
Eredivisie & 507,953  & 406,495 & 49,542 & 51,916 & 79.79\% \\
2022 World Cup & 58,569 & 34,093 & 11,787 & 12,689 & 81.52\% \\
\hline
\end{tabular}
\end{table}

The training of the EPV-OJN model using the Eredivisie dataset employs a cyclic learning rate \citep{smith2017cyclical}, which fluctuates between a base learning rate of $1 \times 10^{-6}$ and a maximum learning rate of $1 \times 10^{-4}$ following a triangular policy with a full cycle lasting 8 epochs. This method helps to avoid local minima. Subsequently, we fine-tune the model using data from the 2022 World Cup, where the maximum learning rate is decreased to $1 \times 10^{-5}$.

A batch size of 128 is used for all EPV-OJN models. This size helps improve training speed and accuracy for pass value models, as larger batches increase the likelihood of including goal-scoring events. Training stops when the validation loss does not improve for 8 consecutive epochs. After training converges, we select the epoch that provides a suitable balance between loss and calibration (as measured by ECE). One epoch covers the complete training set, and one validation step covers all validation samples. The Adam optimizer \citep{Kingma2014AdamAM} with default settings in TensorFlow 2.18 \citep{tensorflow2015-whitepaper} is employed for all models.

The models are trained and evaluated on the basis of their respective loss functions. For the pass likelihood model, binary cross-entropy loss is used to compare the predicted and actual pass destinations. The pass success model also uses binary cross-entropy loss to compare predicted pass success probabilities against actual outcomes.

For the pass value model, categorical cross-entropy loss is applied to the output probabilities (e.g., [0.1, 0.8, 0.1] indicating the probabilities of scoring a goal, having no goal, and conceding a goal within 15 seconds, respectively) compared to ground truth (e.g., [0, 1, 0] if no goal is scored within 15 seconds). The 15-second duration is based on \cite{Fernández2021}, reflecting the average possession duration in football.

Both pass success and pass value models employ temperature scaling as a post-processing step. The optimal temperature value, ranging from 0.1 to 2 with a step size of 0.1, is selected to minimize the calibration error on the validation set. Model calibration is measured using Expected Calibration Error (ECE) as described in \cite{Fernández2021}, pushing the predicted probabilities to align with the actual probabilities.

 \section{Results and Interpretation}
This section presents our main results and findings, beginning with a feature ablation study. Building upon the best feature set identified, we then present the results of an architecture study to find the best number of parameters for OJN-EPV. The best-performing architecture is next used to evaluate the performance of the resulting model. We highlight the inclusion of ball height as a key model feature, demonstrating its impact on enhancing game state predictions. Finally, we conduct a quantitative evaluation of the best-performing OJN-EPV model using the OJN-Pass-EPV benchmark.

\subsection{Feature Ablation Study}\label{ablation_study}

In our feature ablation study, we investigate the effects of various feature combinations on model performance. Across all models, adding features beyond the \textit{fundamental features} (i) player positions and velocities, (ii) distance to the ball for every location, and (iii) ball height, yielded only negligible performance gains in aggregate metrics such as loss and ECE. However, closer inspection of the value models, including visual evaluations of predicted probability surfaces, reveals that \textit{distance to goal} and \textit{angle to goal} substantially improve contextual accuracy. Without these features, the model underestimates value in scenarios near the penalty area.

We hypothesize that the lack of improvement in overall loss and ECE metrics is partly due to the relatively small number of goals in football, causing these effects to be overshadowed. However, retaining distance to goal and angle to goal produces more realistic and domain-consistent predictions, underscoring the importance of complementing quantitative measurements with qualitative assessments when refining EPV models.

\subsection{Architecture Study}

To validate the model architecture, we examine various setups, specifically focusing on the number of filters (8, 16, 32) and the filter dimensions (3x3, 5x5). We find that using only 8 filters significantly reduces performance for all models except the pass value models. We hypothesize that the outputs of the value models are relatively simple and consistent across different game states, primarily because the predicted value is largely influenced by the position of the successful or failed pass. Conversely, the success and likelihood models have more intricate outputs that depend heavily on the specific game state, requiring a greater capacity to effectively capture the dynamics of each scenario. For the models using 32 filters, we find that these models do show slightly better performance compared to the models using 16 filters, but due to only slight improvements and significantly more parameters, especially in the case of 32 filters and a filter dimension of 5x5, we choose to use 16 filters with dimension 5x5 for the OJN-EPV model. This configuration results in 372.355 parameters for the pass success model, 372.359 for the pass likelihood model, and 373.201 parameters for both pass value models (successful and unsuccessful). Note that this is the model described in Section \ref{model_architecture_and_selection}

\subsection{Model Performance}
This section assesses the performance of the OJN-EPV model, focusing on the loss and calibration metrics.

We begin by presenting the loss results for the success, likelihood, and value models. The value models, in particular, showcase the new computation of the loss. This is derived from the probabilities obtained from the softmax function for each class (-1 (conceding a goal) , 0 (no goal), and 1 (scoring a goal)). We use categorical cross entropy loss for the value model and binary cross-entropy loss for the per-class loss computation. To get the overall ECE for the value models, we subtract the probability that a goal is scored by the probability that a goal is conceded.

We use ECE with 10 bins as the calibration metric. To enhance model calibration, temperature scaling is applied to the pass success and pass value models. All models, except one, show the best calibration with a temperature of 1.0, thus no temperature scaling. Only the value unsuccessful model trained and validated on the Eredivisie data shows better calibration with a temperature value of 1.1.

Table \ref{tab:overall_model_performance_test_eredivisie} displays the various models trained solely on Eredivisie data and Table \ref{tab:overall_model_performance_test_world_cup} the models trained additionally on World Cup data, where training begins with Eredivisie model weights and fine-tuning is done using the World Cup dataset. All models are tested on both datasets. It is noted that the success and likelihood models show reduced loss values when assessed on the World Cup dataset, as opposed to the Eredivisie dataset. This improvement likely stems from multiple factors: the competition level in World Cup matches may be more uniform, and the World Cup data could be of higher quality. Furthermore, we notice that fine-tuning on the World Cup dataset does not improve loss measures nor calibration measures significantly.

\begin{table}[h!]
\centering
\caption{Loss and ECE on both datasets for models trained on Eredivisie data}
\label{tab:overall_model_performance_test_eredivisie}
\scriptsize

\vspace{5mm}
\begin{tabular}{|l|p{1.8cm}|p{1.8cm}|p{1.8cm}|p{1.8cm}|p{1.8cm}|}
\hline
\textbf{Model} & \textbf{Loss on Eredivisie} & \textbf{ECE on Eredivisie} & \textbf{Loss on World Cup} & \textbf{ECE on World Cup} \\ \hline
Pass Success & 0.1558 & 0.0024 & 0.1355 & 0.0122\\ \hline
Pass Likelihood & 4.7225 & - & 4.4528 & - \\ \hline
Pass Value (Successful) & 0.0689 & 0.0016 & 0.0835 & 0.0060 \\ \hline
Pass Value (Unsuccessful) & 0.0663 & 0.0042 & 0.0726 & 0.0056\\ \hline
\end{tabular}
\end{table}

\begin{table}[h!]
\centering
\caption{Loss and ECE on both datasets for models fine-tuned on World Cup data}
\label{tab:overall_model_performance_test_world_cup}
\scriptsize

\vspace{3mm}
\begin{tabular}{|l|p{1.8cm}|p{1.8cm}|p{1.8cm}|p{1.8cm}|p{1.8cm}|}
\hline
\textbf{Model} & \textbf{Loss on Eredivisie} & \textbf{ECE on Eredivisie} & \textbf{Loss on World Cup} & \textbf{ECE on World Cup} \\ \hline
Pass Success & 0.1568 & 0.0090 & 0.1326 & 0.0047 \\ \hline
Pass Likelihood & 4.7227 & - & 4.4367 & - \\ \hline
Pass Value (Successful) & 0.0687 & 0.0024  & 0.0836 & 0.0065  \\ \hline
Pass Value (Unsuccessful) & 0.0671 & 0.0045 & 0.0740 & 0.0050 \\ \hline
\end{tabular}
\end{table}

The detailed loss per class based on the value model, as shown in Tables \ref{tab:detailed_loss_by_class_test_eredivisie} and \ref{tab:detailed_loss_by_class_test_world_cup}, provides a representation of the model's subtle understanding of pass outcomes. Low loss and calibration scores indicate the model's capability in precisely forecasting if a pass will lead to scoring, conceding, or no goal within 15 seconds for the team in ball possession.

\begin{table}[h!]
\centering
\caption{Pass value model - loss and ECE by class for model trained on Eredivisie data}
\label{tab:detailed_loss_by_class_test_eredivisie}
\footnotesize

\vspace{3mm}
\begin{tabular}{|l|c|c|c|c|}
\hline
\multicolumn{5}{|c|}{\textbf{Pass Value Model - Detailed Loss by Class}} \\ \hline
\textbf{Class} & \textbf{Loss on Eredivisie} & \textbf{ECE on Eredivisie} & \textbf{Loss on World Cup} & \textbf{ECE on World Cup} \\ \hline
Scoring goal (Successful Passes) & 0.0590 & 0.0018 & 0.0689 & 0.0048 \\ \hline
No goal (Successful Passes) & 0.0660 & 0.0040 & 0.0791 & 0.0035 \\ \hline
Conceding goal (Successful Passes) & 0.0096 & 0.0023 & 0.0144 & 0.0022 \\ \hline
Scoring goal (Unsuccessful Passes) & 0.0379 & 0.0060 & 0.0357 & 0.0074 \\ \hline
No goal (Unsuccessful Passes) & 0.0620 & 0.0105 & 0.0663 & 0.0134 \\ \hline
Conceding goal (Unsuccessful Passes) & 0.0271 & 0.0041 & 0.0362 & 0.0062 \\ \hline
\end{tabular}
\end{table}

\begin{table}[h!]
\centering
\caption{Pass value model - loss and ECE by class for model fine-tuned on World Cup data}
\label{tab:detailed_loss_by_class_test_world_cup}
\footnotesize

\vspace{3mm}
\begin{tabular}{|l|c|c|c|c|}
\hline
\multicolumn{5}{|c|}{\textbf{Pass Value Model - Detailed Loss by Class}} \\ \hline
\textbf{Class} & \textbf{Loss on Eredivisie} & \textbf{ECE on Eredivisie} & \textbf{Loss on World Cup} & \textbf{ECE on World Cup} \\ \hline
Scoring goal (Successful Passes) & 0.0590 & 0.0011 & 0.0693 & 0.0057 \\ \hline
No goal (Successful Passes) & 0.0658 & 0.0032 & 0.0793 & 0.0042 \\ \hline
Conceding goal (Successful Passes) & 0.0093 &0.0021 & 0.0141 &0.0020 \\ \hline
Scoring goal (Unsuccessful Passes) & 0.0385 & 0.0064 & 0.0366 & 0.0081 \\ \hline
No goal (Unsuccessful Passes) & 0.0629 & 0.0110 & 0.0675 & 0.0155 \\ \hline
Conceding goal (Unsuccessful Passes) & 0.0273 & 0.0046 & 0.0366 & 0.0071 \\ \hline
\end{tabular}
\end{table}

\subsection{Influence of Ball Height on Predictions}
\begin{figure}[h!]
\centering
\begin{subfigure}{0.48\textwidth}
    \includegraphics[width=\linewidth]{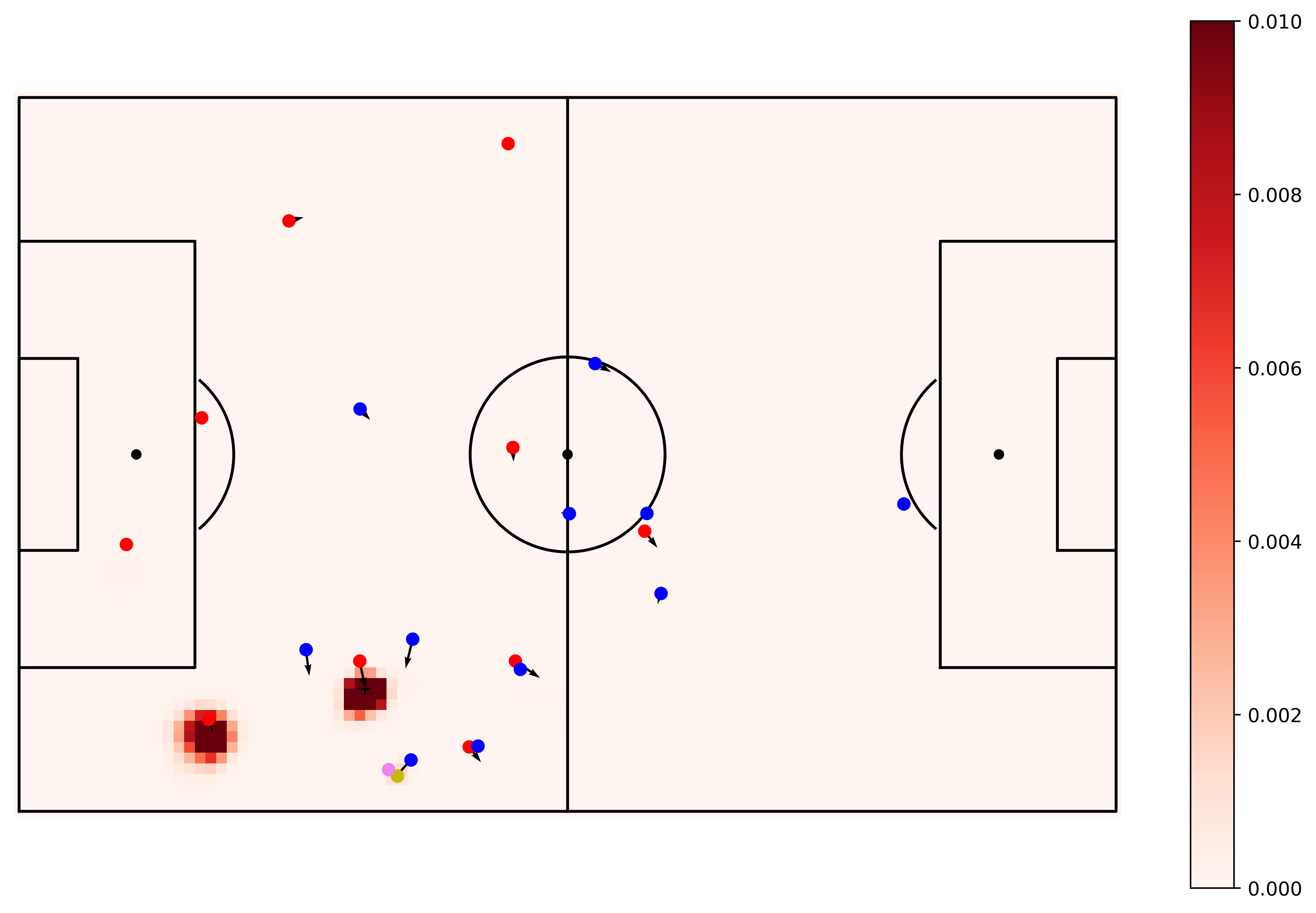}
    \caption{Ground pass scenario (0 meters ball height)}
    \label{fig:ball_height_ground}
\end{subfigure}
\hfill
\begin{subfigure}{0.48\textwidth}
    \includegraphics[width=\linewidth]{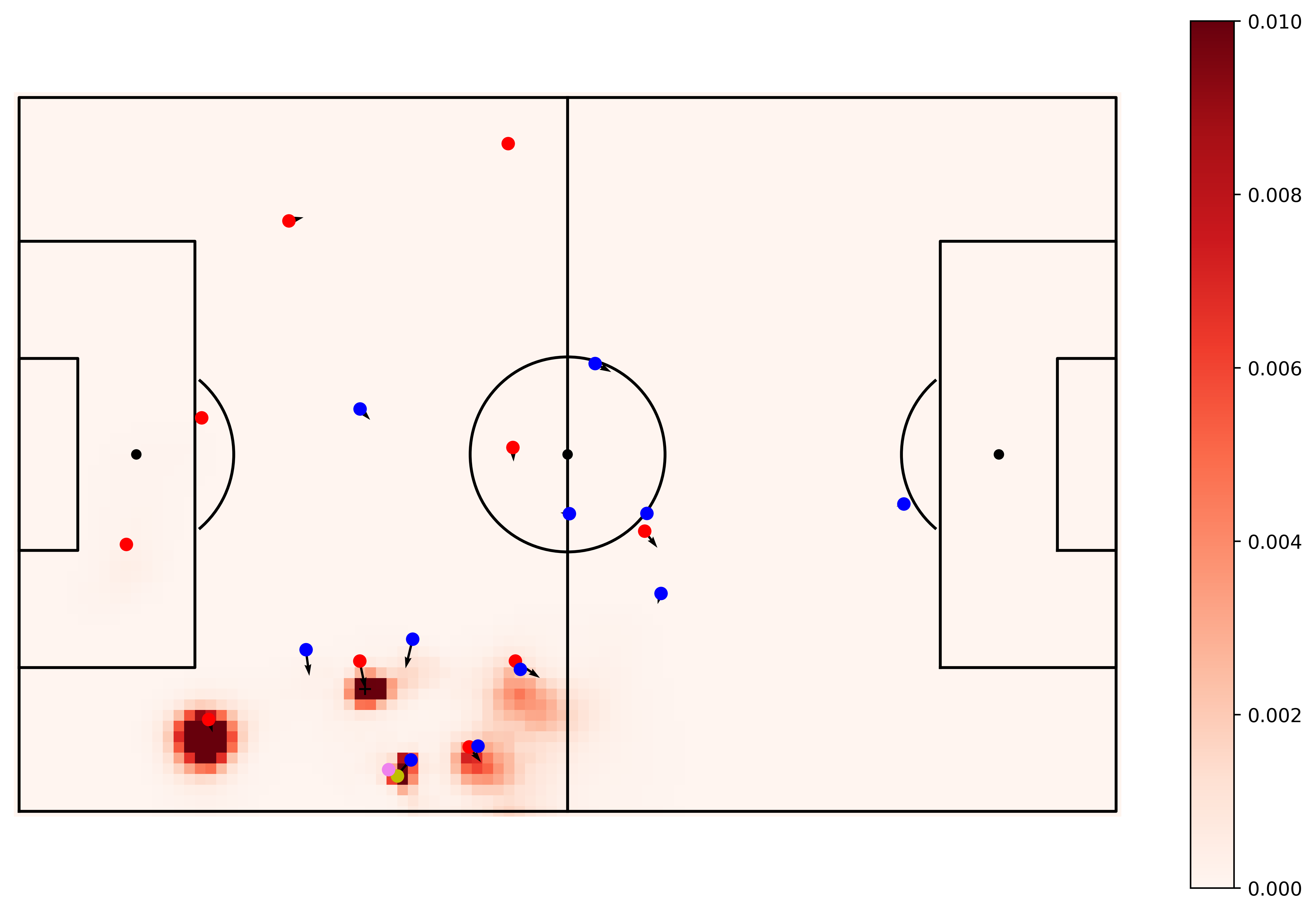}
    \caption{Aerial pass scenario (2 meters ball height)}
    \label{fig:ball_height_aerial}
\end{subfigure}
\caption{Comparative visualization of pass likelihood based on ball height. In both scenarios, the red team (red dots) is in possession, with the ball carrier highlighted in yellow and the ball depicted in violet. The opposing team is represented in blue (blue dots). The color intensity on the pitch indicates the likelihood of the pass reaching a location - the darker the shade, the higher the probability. Figure~\ref{fig:ball_height_aerial} shows how the model recognizes that aerial passes can be made over opponents, while also showing increased uncertainty about the pass destination compared to the ground pass in Figure~\ref{fig:ball_height_ground}.}
\label{fig:ball_heights}
\end{figure}

Incorporating the ball's height (z-axis) into the model significantly impacts the pass likelihood predictions by adding the crucial vertical dimension to the analysis. Figure \ref{fig:ball_heights} illustrates this by contrasting two scenarios: a ground pass \ref{fig:ball_height_ground} and an aerial pass with the ball at 2 meters high \ref{fig:ball_height_aerial}. In the aerial pass scenario, the model recognizes that the ball can be passed over opponents and also shows increased uncertainty about the pass's destination, assuming it will likely be a header. These predictions align with real-world football knowledge: headers are generally less precise than ground passes due to reduced control. While ball height doesn't substantially affect overall loss or ECE metrics because of the rarity of aerial passes, Figure \ref{fig:ball_heights} demonstrates a specific instance where it's highly relevant. This highlights the practical importance of considering ball height for specific passing situations. 

\newpage
\subsection{Relevance of Decomposing Pass Value in Reward and Risk}

\begin{figure}[h!]
\centering
\includegraphics[width=\linewidth]{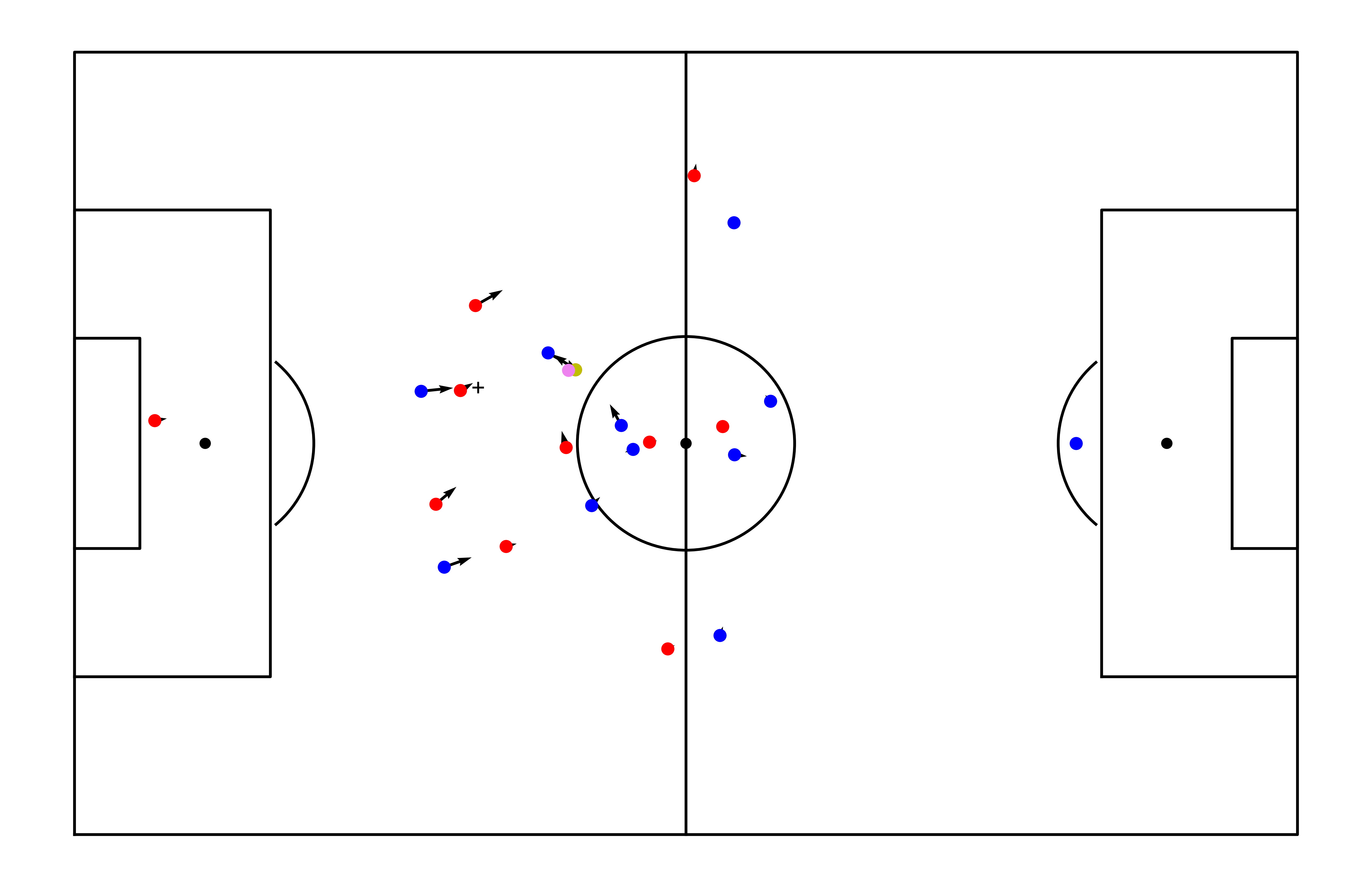}
\caption{Game state showing a pass scenario. The ball carrier, marked with a yellow dot and belonging to the red team, is making a pass to the target location marked by "+".}
\label{fig:event1500}
\end{figure}

Decomposing pass EPV into reward and risk components offers a more nuanced perspective on the complexities of passing decisions in football. By separating the potential positive and negative impacts of a pass, we can gain deeper insights into the underlying volatility of seemingly straightforward game states.  Figure \ref{fig:event1500} depicts one such game state, where the decomposition of EPV reveals the trade-offs inherent in a passing decision.

In this game state, the pass value model trained on successful passes assigns a slightly negative overall value (i.e., -0.0047) for the end location of the pass marked with a "+". This implies that even if the pass is completed, the blue team is still considered more likely to score within 15 seconds than the red team. Unlike earlier approaches that define risk as the value associated with losing possession and reward as the value associated with keeping the ball, our analysis highlights a different perspective. Even a successful pass can lead to an unpredictable game state, potentially detrimental for the team in possession, depending on factors such as opponent pressure. This assessment is based on potential outcomes: the blue team has a 0.0199 probability of scoring compared to 0.0152 for the red team. This analysis illustrates the complex trade-offs that can be present in passing decisions.

\newpage

\subsection{Benchmark performance}\label{benchmark_performance}

Prior to incorporating \emph{distance to goal} and \emph{angle to goal} features in our pass value models, the benchmark performance for the model trained only on the Eredivisie dataset was 68\% on the OJN-Pass-EPV benchmark, rising to 70\% after fine-tuning on the World Cup data. Once we added these two features, the global loss and calibration metrics did not show a pronounced improvement (see Section~\ref{ablation_study}). However, the benchmark performance for both the Eredivisie-trained and World Cup fine-tuned models increased to 78\%.

This result underscores the importance of supplementing aggregate metrics (like loss and calibration) with a real-world evaluation such as our OJN-Pass-EPV benchmark. A pair of features (distance and angle to goal) can make the difference in identifying the game state with a higher probability of scoring even if the global loss and ECE measures remain largely unchanged.

The OJN-EPV model thus achieves a performance level of 78\% on our OJN-Pass-EPV benchmark. The benchmark creation, as described in Section \ref{benchmark_creation}, was designed to include a considerable number of challenging game state pairs, such as those with small but impactful differences in player positioning. Even a 0.5-meter shift in position can influence the value of a game state, and with a model granularity of approximately 1 x 1 meter, this poses a clear challenge.

One conclusion is that the OJN-EPV model underperforms in scenarios involving offside players, assigning a higher pass EPV to a game state in which a player is offside (resulting in an inevitable unsuccessful pass). This issue arises in three instances within the benchmark. Other pairs also highlight scenarios where the model fails to recognize the more valuable game state, reflecting the benchmark’s intent to include difficult cases that guide further model improvements. Nonetheless, with a benchmark performance of 78\%, the OJN-EPV model generally delivers accurate predictions across a wide range of game state pairs.

 \section{Discussion}

This research contributes to the advancement of EPV models for football by introducing the first EPV benchmark dataset and by validating an enhanced EPV model across distinct competitions, namely the Dutch Eredivisie and the 2022 World Cup. The novel integration of ball height as a feature and the segmentation of pass value into distinct reward and risk components refine the model's predictive capabilities and augment its interpretability. These enhancements enable a more nuanced analysis of pass decisions, facilitating a deeper understanding of their potential outcomes.

\begin{figure}[h]
\centering
\includegraphics[width=1.0\linewidth]{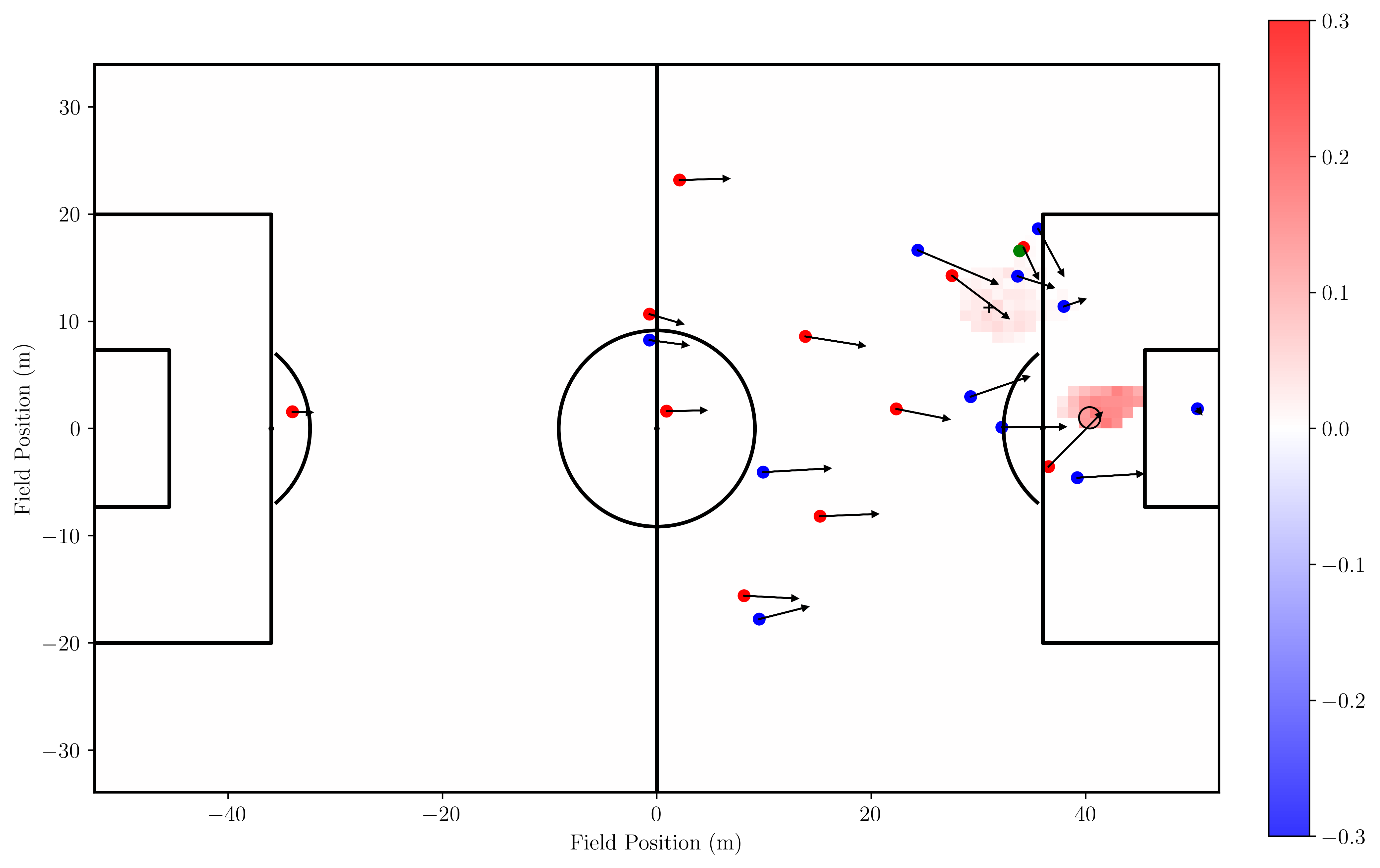}
\caption{Analysis of pass decisions with OJN-EPV (based on the output in Definition~\ref{def:output}). This figure shows a player's actual pass (marked by "+") against the optimal pass location (circled) as identified by OJN-EPV. The comparison highlights potential areas for decision-making refinement, illustrating how OJN-EPV can assist in identifying more dangerous pass options.}
\label{fig:optimal_value_location}
\end{figure}

In application, the model could serve as an aid in optimizing pass strategies by contrasting players' actual choices with theoretically optimal ones, as illustrated in Figure \ref{fig:optimal_value_location}. This comparison not only identifies opportunities for improvement but also highlights the importance of strategic pass selection in influencing game outcomes. 

\newpage

\newtheorem{definition}{Definition}
\small
\begin{definition}\label{def:output}
\begin{gather*}
\mathrm{Output}(x, y) = \begin{cases} V(x, y) & \text{if } L(x, y) > 0.001 \\ 0 & \text{otherwise} \end{cases} \\
\text{where } V(x, y) = S(x, y) V_{s}(x, y) + (1 - S(x, y)) V_{u}(x, y) \\
V_s(x, y) = P_{\text{score}}(x, y | \text{success}) - P_{\text{concede}}(x, y | \text{success}) \\
V_u(x, y) = P_{\text{score}}(x, y | \text{no success}) - P_{\text{concede}}(x, y | \text{no success}) \\
\begin{aligned}
\text{and } & V(x,y) : \text{estimated value of a pass that ends up at location } (x, y),\\
& L(x, y) : \text{likelihood that a pass ends up at location } (x, y), \\
& S(x, y) : \text{probability of a successful pass to} (x, y), \\
& P_{\text{score}}(x, y | \text{success}) : \text{probability of scoring after a successful pass to } (x, y), \\
& P_{\text{concede}}(x, y | \text{success}) : \text{probability of conceding after a successful pass to } (x, y), \\
& P_{\text{score}}(x, y | \text{no success}) : \text{probability of scoring after an unsuccessful pass to } (x, y), \\
& P_{\text{concede}}(x, y | \text{no success}) : \text{probability of conceding after an unsuccessful pass to } (x, y), \\
\end{aligned}
\end{gather*}
\end{definition}

Definition~\ref{def:output} underscores how OJN-EPV can be applied to focus on likely pass destinations (those with $L(x, y) > 0.001$) before computing the value of a potential pass. By emphasizing a threshold that captures sufficiently probable pass locations, we concentrate our attention on meaningful pass options. 

While this work advances the state of EPV modeling, additional steps are required for broad practical adoption. For instance, OJN-EPV does not explicitly capture the intended recipient of a pass, and integrating that intention would help differentiate the quality of a decision from the quality of its execution \citep{Power2017, peralta2020seeing, dick2022can, spearman2017physics}. Including more extensive datasets, adding player-specific properties, and assessing EPV throughout entire matches rather than discrete events would further enrich the model. These efforts would facilitate an even deeper understanding of both individual and team performance, reinforcing the utility of EPV in guiding tactical strategies and player evaluations.

 \section{Conclusion}
In this paper, we introduce the OJN-Pass-EPV benchmark of game state pairs with relative EPVs, which allows to quantitatively evaluate pass EPV models and provides a template to evaluate overall EPV models and all their components. We also introduce the OJN-EPV model, which has improved performance by incorporating the z-axis of the ball, demonstrates robust performance across a new dataset, and enables a more granular interpretation of passes by dividing the pass value model into reward and risk components. Incorporating ball height and risk-reward decomposition delivers more accurate and insightful evaluations of passing decisions, ultimately fostering a more data-driven and strategic approach to the game.
\newpage

\end{document}